\documentclass[review]{elsarticle}

\usepackage{lineno,hyperref}
\modulolinenumbers[5]

\usepackage{longtable} %per multipage table
\usepackage{xcolor}
\usepackage{amsmath}
\usepackage{booktabs}

\usepackage{caption}
\usepackage{subcaption}
\usepackage{multirow}

\journal{Journal of Expert Systems with Applications}

\usepackage{etoolbox}
\patchcmd{\pprintMaketitle}
{\ifvoid\absbox\else\unvbox\absbox\par\vskip10pt\fi}
{\ifvoid\absbox\else\clearpage\unvbox\absbox\par\vskip30pt\fi}
{}{}
\patchcmd{\pprintMaketitle}
{\hrule\vskip12pt}
{}
{}{}
\patchcmd{\pprintMaketitle}
{\hrule\vskip12pt}
{}
{}{}
\appto{\pprintMaketitle}{\clearpage}
\biboptions{authoryear}

\begin{document}
	
	\begin{frontmatter}
		
		\title{Genetic algorithm for feature selection of EEG heterogeneous data}
		%\tnotetext[mytitlenote]{Fully documented templates are available in the elsarticle package on \href{http://www.ctan.org/tex-archive/macros/latex/contrib/elsarticle}{CTAN}.}
		
		%% Group authors per affiliation:
		\author{Aurora Saibene\corref{cor1}}
		\ead{a.saibene2@campus.unimib.it}
		
		\cortext[cor1]{Corresponding author}
		
		\author{Francesca Gasparini}
		\ead{francesca.gasparini@unimib.it}
		
		\address{University of Milano-Bicocca, Department of Informatics, Systems and Communications, Multi Media Signal Processing Laboratory, Viale Sarca 336, 20126, Milan, Italy}
		\address{University of Milano-Bicocca, NeuroMI, Piazza dell’Ateneo Nuovo 1,
			20126, Milan, Italy}
		
		%A concise and factual abstract is required. The abstract should state briefly the purpose of the research, the principal results and major conclusions. An abstract is often presented separately from the article, so it must be able to stand alone. For this reason, References should be avoided, but if essential, then cite the author(s) and year(s). Also, non-standard or uncommon abbreviations should be avoided, but if essential they must be defined at their first mention in the abstract itself.
		\begin{abstract}
		The electroencephalographic (EEG) signals provide highly informative data on brain activities and functions. However, their heterogeneity and high dimensionality may represent an obstacle for their interpretation. The introduction of a priori knowledge seems the best option to mitigate high dimensionality problems, but could lose some information and patterns present in the data, while data heterogeneity remains an open issue that often makes generalization difficult. In this study, we propose a genetic algorithm (GA) for feature selection that can be used with a supervised or unsupervised approach. Our proposal considers three different fitness functions without relying on expert knowledge. Starting from two publicly available datasets on cognitive workload and motor movement/imagery, the EEG signals are processed, normalized and their features computed in the time, frequency and time-frequency domains. The feature vector selection is performed by applying our GA proposal and compared with two benchmarking techniques.
		The results show that different combinations of our proposal achieve better results in respect to the benchmark in terms of overall performance and feature reduction. Moreover, the proposed GA, based on a novel fitness function here presented, outperforms the benchmark when the two different datasets considered are merged together, showing the effectiveness of our proposal on heterogeneous data.
			
		\end{abstract}
		
		
	\end{frontmatter}
	
	%\linenumbers
	
	% % % % % NUOVA STRUTTURA % % % % %
	
	% % % % % INTRODUCTION % % % % %
	\section{Introduction}\label{sec:intro}
	% State the objectives of the work and provide an adequate background, avoiding a detailed literature survey or a summary of the results.
	Electroencephalography (EEG) has been employed by researchers in different experimental domains, regarding emotion recognition, motor imagery, mental workload, seizure detection and sleep \citep{craik2019deep}, mainly because it is non-invasive, has recently been embedded in low-cost devices \citep{larocco2020systemic}, produces a signal with high temporal resolution \citep{rojas2018study}, and its data bring intrinsic brain information.\\   
	In fact, the EEG signal corresponds to a single subject's brain electric potentials, recorded in a certain period of time through electrodes placed on his/her scalp \citep{hosseini2020review} and is characterized by different frequency bands (or rhythms).  These rhythms are representative of specific brain activities \citep{vaid2015eeg}: the delta ($\delta \leq 4$ Hz) frequency band is present during deep sleep, while the theta ($\theta = [4 - 8]$ Hz) rhythm is elicited by emotional stress and is present in adults that are drowsy or sleeping; the alpha ($\alpha = [8 - 13]$ Hz) frequency band represents a relaxed state while being aware, the beta ($\beta = [13 - 30]$ Hz) rhythm is typical of alertness states that may involve active thinking and attention, and the gamma ($\gamma \geq 31$ Hz) rhythm is elicited by intensive brain activities during consciousness.\\
	Knowing the EEG signal characteristics, many features of different types in the time, frequency and time-frequency domains can be extracted from these data \citep{jenke2014feature, boonyakitanont2020review} in order to discriminate a variety of EEG experimental or natural conditions. However, this may bring to high dimensional data that may present redundant and noisy features \citep{cimpanu2017multi}.\\
	Therefore, reducing the feature vector is of utmost importance to increase the possibility of finding data patterns and better representations, while improving model efficiency, without incurring in curse of dimensionality and overfitting \citep{nakisa2018evolutionary, lotte2018review}.\\
	Most of the researchers tend to perform a manual selection of the features or to restrict their analyses to a specific domain, to use dimensionality reduction techniques, especially the Independent Component Analysis (ICA) and the Principal Component Analysis (PCA) \citep{nicolas2012brain}, strongly influencing the results with expert knowledge.\\
	However, notice that the EEG signals are heterogeneous due to their variability in time, being influenced by experimental protocols and environmental conditions, varying between subjects and even in a same subject \citep{roy2019deep}. Therefore, the feature selection problem may be addressed by exploiting techniques that rely on dynamic probability optimization \citep{atyabi2012evolutionary}.\\
	In fact, many researches have focused on the application of evolutionary algorithms for feature selection in a specific EEG experimental domain: from emotion recognition \citep{shon2018emotional, nakisa2018evolutionary} to motor imagery \citep{baig2017differential, leon2019feature, saibene2020centric}, from biometric systems \citep{moctezuma2020towards} to epilepsy \citep{guo2011automatic, wen2017effective}.\\
	Especially, one of the most employed evolutionary techniques among the ones proposed by different authors, is the genetic algorithm (GA) \citep{goldberg1988genetic} with its variations \citep{rejer2013genetic, amarasinghe2016eeg, wen2017effective,rejer2018gamers, shon2018emotional,leon2019feature, moctezuma2020towards}, due to the fact that it has many appealing characteristics. Apart from being able to perform the feature selection without relying on expert knowledge and thus avoiding the introduction of assumptions on the features, as other evolutionary algorithms \citep{xue2015survey}, it is robust and efficient \citep{shon2018emotional}. Moreover, it is versatile, being its configuration easy to modify \citep{babatunde2014genetic}.\\
	From a brief literature review, we confirmed that GA variations are widely used and employ a binary coding for the EEG feature vector \citep{kolodziej2011new, amarasinghe2016eeg, cimpanu2017multi, wen2017effective, leon2019feature, moctezuma2020towards}. The authors also prefer the usage of supervised learning techniques, e.g., Linear Discriminant Analysis (LDA) \citep{kolodziej2011new} and Support Vector Machines (SVMs) \citep{rejer2013genetic, amarasinghe2016eeg, cimpanu2017multi, leon2019feature}, and external measures to evaluate both fitness functions and models. The fitness functions mainly involve classification error \citep{guo2011automatic, kolodziej2011new} or a combination of false positive and true positive rate \citep{wen2017effective} minimization, accuracy maximization \citep{rejer2013genetic, amarasinghe2016eeg} that may be considered in a multi-objective approach with the optimization of the selected feature number \citep{cimpanu2017multi, saibene2020centric}.\\
	There are just a couple of works using an unsupervised learning approach to complement their GA feature selection. Some unsupervised strategies are used only to cluster the data while taking track of a cluster centroids and objects \citep{rejer2018gamers} or to achieve the best inter-cluster separability \citep{shon2018emotional}, despite using again an external measure (accuracy) for evaluation. These works seem to never employ internal measures in their fitness functions or in the final evaluation of their results, and avoid the usage of standard clustering techniques like K-means clustering. In fact, to the best of our knowledge, K-means clustering has not been used as part of GAs for EEG feature vector reduction, even though it has been presented in different works regarding the EEG signal decoding \citep{liang2019unsupervised}, class definition \citep{gurudath2014drowsy}, pattern detection \citep{kabir2018computer}, clustering \citep{alkan2011use, asanza2017eeg}.\\
	Also notice that, even though the evolutionary algorithms for feature selection are computationally demanding, especially when using a wrapper approach \citep{xue2018novel}, a restricted number of studies \citep{leon2019feature} have shown interest in the time required for their computation. To address this issue, most of the EEG related works restrict their analyses to a subset of the original data \citep{nakisa2018evolutionary} or on datasets with a narrow number of subjects (from a minimum of 1 to a maximum of 8 subjects) \citep{baig2017differential, zhao2018analyze, rejer2018gamers, leon2019feature}. Moreover, some researchers have introduced a maximum number of features to select \citep{kolodziej2011new} and all the reported papers present analyses with a subject-based approach.\\
	Motivated by the previous studies, in this work we propose a modified GA for feature selection as an evolution of the one presented in our previous study \citep{saibene2020centric}, to promote the usage of different machine learning approaches and fitness functions on heterogeneous datasets. Therefore, the contributions given by our proposal are:
	\begin{itemize}
		\item The introduction of a normalization procedure to mitigate the EEG signal heterogeneity and thus allow a population-based analysis, i.e., on a group of subjects instead of being performed on a single subject, the mixing of very different datasets and the possible generalizability of the proposal;
		\item The computation of a variety of features in the time, frequency and time-frequency domain performed on different electrodes and rhythms, to avoid introducing expert knowledge into the analyses;
		\item A GA for feature selection that may be used with (i) a supervised learning approach by employing a linear SVM and with an unsupervised one by using a K-means clustering algorithm; (ii) different fitness functions that may consider only the performance measures, i.e., accuracy for the supervised model and silhouette for the unsupervised one, or a combination of these measures and the number of features, weighted by specific parameters; (iii) a set of stopping criteria that takes into consideration performance and time constraints and that may dynamically affect the total number of generations.
	\end{itemize}   
	The basic idea of our approach is to provide a generalizable feature selection that (i) avoids an excessive computational time, (ii) is flexible, being tested with different parameters and conditions, (iii) returns a feature vector that allows a better data representation in respect to the benchmarking results. The benchmark is obtained by computing the machine learning models on the datasets retaining all the features and the datasets reduced using PCA.\\
	Therefore, the paper is organized as follows. In Section \ref{sec:ourproposal}, a brief overview of our proposal is given. In Section \ref{sec:materialdescription}, the materials employed for our approach testing and the details on the computed features are described. Subsequently, the GA feature selection procedure is detailed (Section \ref{sec:eeg_feature_selection}).  The results obtained by testing our approach with different experiments are reported and discussed in Section \ref{sec:results}. Section \ref{sec:conclusions} summarizes the contributions and limitations of the presented work, while proposing some future developments for GA feature selection.
	
	% % % % % OUR PROPOSAL % % % % %
	\section{Our proposal}\label{sec:ourproposal}
\begin{figure}[htb!]
		\centering
		\includegraphics[width=\textwidth]{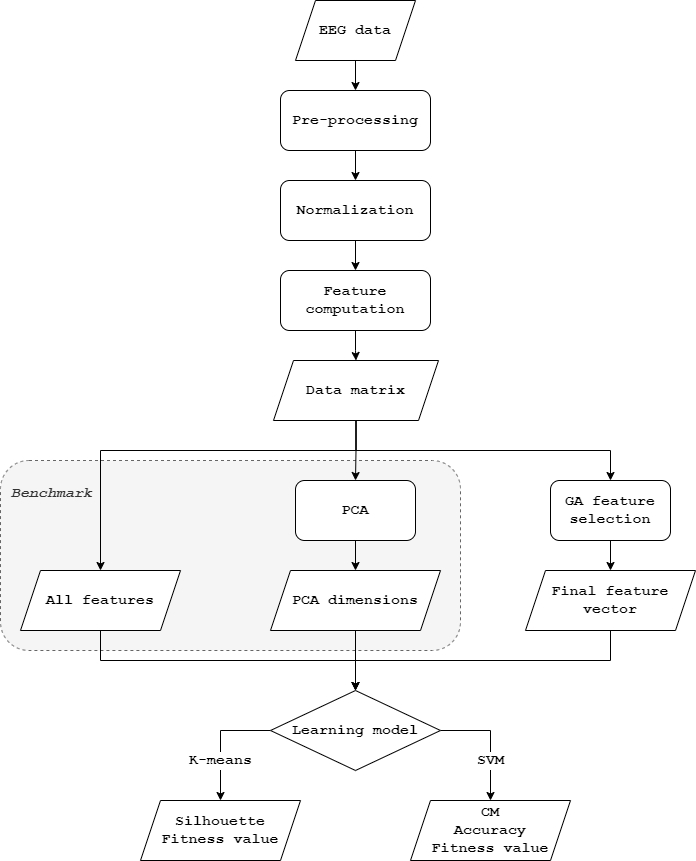}
		\caption{Flowchart of our proposed approach.}
		\label{fig:our_proposal_fc}
	\end{figure}

	The main steps of the proposed approach are depicted in the flowchart in Fig. \ref{fig:our_proposal_fc}. The core procedures, i.e., feature computation and GA for feature selection, are described in details in dedicated sections (Section \ref{sec:eeg_feature_computation} and \ref{sec:eeg_feature_selection}).\\	
	The first computational step is represented by the EEG signal pre-processing to reduce the noise affecting the data, if needed. It consists of a combination of high-pass and low-pass Finite Impulse Response (FIR) filters \citep{proakis2004digital}, followed by a notch filter \citep{proakis2004digital} to remove power line noise.\\
	Afterwards, the z-score \citep{patro2015normalization} data normalization is performed to mitigate the heterogeneity present in the EEG signals and allow data comparison \citep{saibene2020centric}.\\
	The processed data are then passed as inputs to the feature computation procedure (Section \ref{sec:eeg_feature_computation}), obtaining a data matrix. The rows of this data matrix correspond to the EEG task instances, while its columns to the features computed on the processed data.\\
	Afterwards, we apply different computations to this data matrix: (i) the data matrix is considered as is, (ii) its dimensions are reduced by applying PCA \citep{abdi2010principal}, and (iii) a subset of its features is selected by applying the proposed GA for feature selection (Section \ref{sec:eeg_feature_selection}). The resulting three data matrices are passed to the learning model of interest (SVM or K-means clustering) to compare the three strategies. Finally, we remind that the first two strategies represent the benchmark.
	
	% % % % % MATERIALS DESCRIPTION % % % % %
	\section{Materials description}\label{sec:materialdescription}
	% Provide sufficient details to allow the work to be reproduced by an independent researcher. Methods that are already published should be summarized, and indicated by a reference. If quoting directly from a previously published method, use quotation marks and also cite the source. Any modifications to existing methods should also be described.
	In this section the two employed EEG signal datasets, i.e., the \textit{EEG During Mental Arithmetic Tasks} dataset \citep{zyma2019electroencephalograms, goldberger2000physiobank}, from now on called dataset A, and the \textit{EEG Motor Movement/Imagery Dataset} \citep{schalk2004bci2000, goldberger2000physiobank}, from now on called dataset B, are described. \\
	The computed features in the time, frequency and time/frequency domains are also detailed in this section.	
	
	% DATASETS
	\subsection{Datasets}\label{sec:datasets}
\begin{figure}[htb!]
		\centering
		\begin{subfigure}[b]{0.45\textwidth}
			\centering
			\caption{}
			\includegraphics[width=\textwidth]{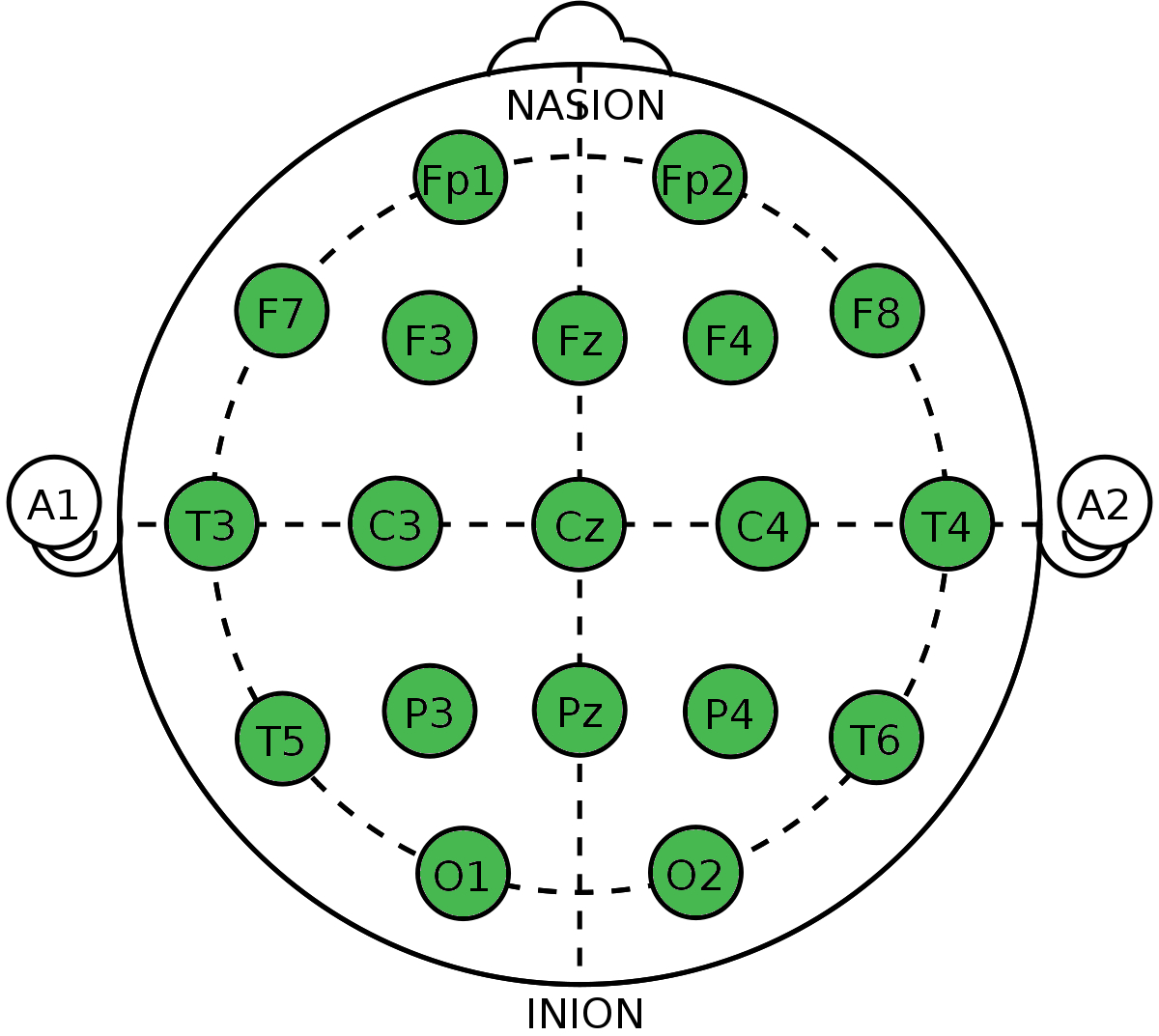}			
			\label{fig:10-20sys-sel}
		\end{subfigure}
		\hfill
		\begin{subfigure}[b]{0.45\textwidth}
			\centering
			\caption{}
			\includegraphics[width=\textwidth]{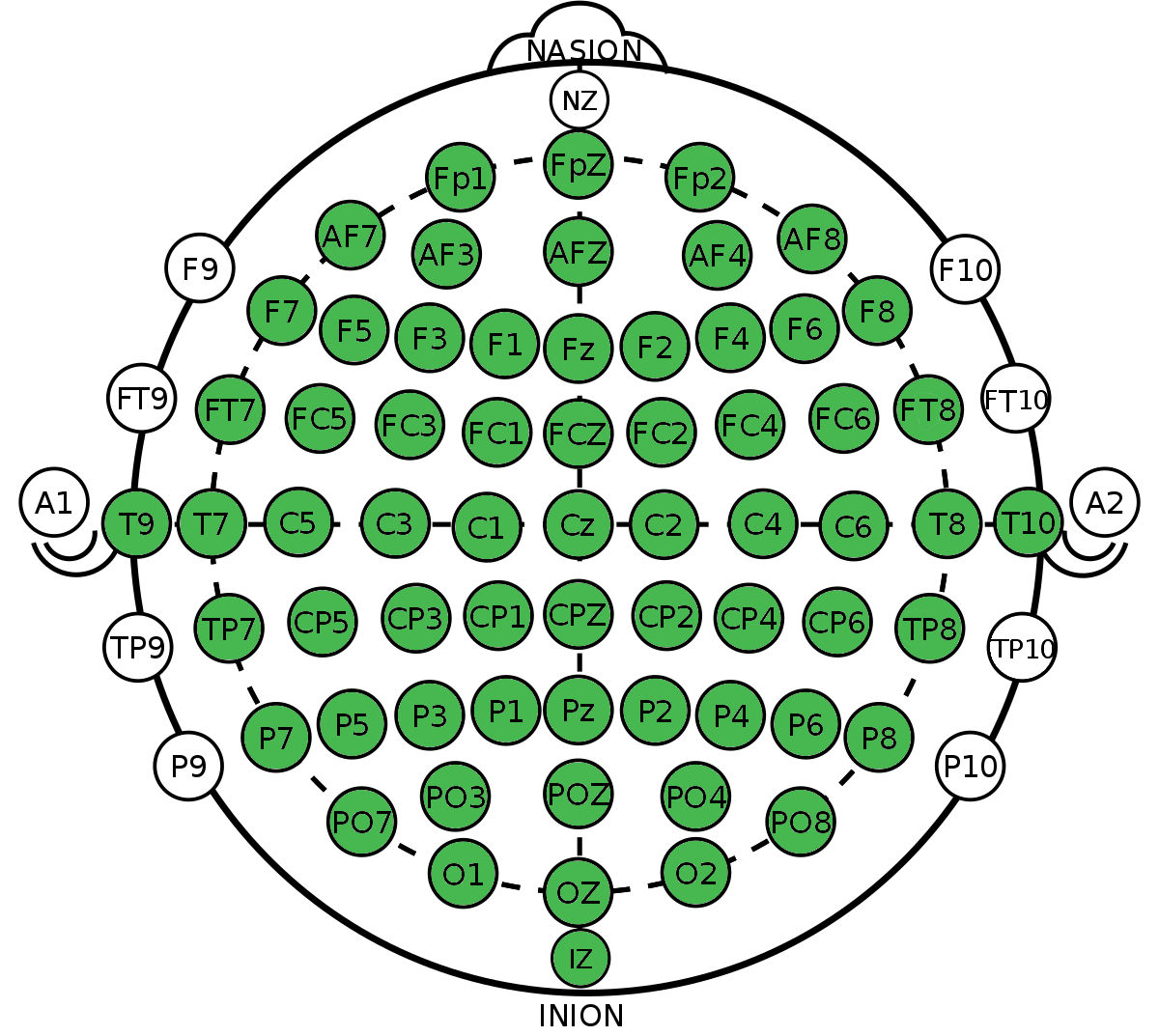}			
			\label{fig:10-10sys-sel}
		\end{subfigure}
		\caption{Electrodes placement on the scalp. The colored electrodes represent the ones employed for the analyses of (a) dataset A and (b) dataset B.}
	\end{figure}

	The choice of dataset A was driven by the presence of two appealing characteristics: (i) the data are pre-processed and analyzed by field experts to remove noise, thus providing an ideal test environment, and (ii) the recording segments correspond to two well defined conditions, labeled by the experimenters.\\
	In fact, dataset A presents for each of the 36 participants (9 males, 27 females with mean age 18.25 and standard deviation 2.17), 2 recordings corresponding to a resting phase (REST) of 180 s and a mental arithmetic task phase (MAT) of 60 s, respectively.\\
	During the MAT, the participants had to undertake an intensive cognitive task and thus \cite{zyma2019electroencephalograms} proposed a sequential arithmetic subtraction.\\
	The EEG signals have been recorded with a sampling rate of 500 Hz and by placing the electrodes on the scalp accordingly to the 10/20 international system. In the present work, the considered electrodes are C$\{3,4,z\}$, F$\{3,4,7,8\}$, Fp$\{1,2\}$, Fz, O$\{1,2\}$, P$\{3,4,z\}$, T$\{3,4,5,6\}$ and are depicted in Fig. \ref{fig:10-20sys-sel}.\\
	Notice that the signal is bandpass filtered in the range $[0.5 - 45]$ Hz and notch filtered at 50 Hz. All the electrodes are referenced to the lobe sensors.\\
	For more details on the experiment, methods and subjects involved, please refer to \cite{zyma2019electroencephalograms}.
		
	\noindent{Wanting to explore the generalizability of our approach, dataset B was introduced having a completely different paradigm in respect to dataset A and presenting a good number of recordings with a variety of experimental conditions.}\\
	109 participants (of unspecified gender and age) were involved in an experiment constituted by 2 subsequent baseline recordings of eyes open and closed, and 4 motor/imagery tasks repeated 3 times:
	\begin{enumerate}
		\item Left and right hand movement;
		\item Left and right hand imagination of movement;
		\item Fists and feet movement;
		\item Fists and feet imagination of movement.
	\end{enumerate}
	The baseline recordings (eyes open or closed) lasted for 4.2 s, while each movement or imagination of movement in the motor/imagery tasks lasted for 4.1 s and was preceded by a rest phase. A specific task was identified by a visual cue.\\
	In this study only the eyes closed (CLOSE) baseline recordings and the first motor/imagery task of left (LH) and right hand (RH) movement are considered. Notice that the motor activity involves the $\alpha$ and $\beta$ frequency bands, especially in the central cortical area \citep{szczuko2018comparison, tang2019classification} and thus the identified condition should be easily detected when the movement is well done and the signal does not present a great amount of noise. \\
	The EEG signals have been recorded with a sampling rate of 160 Hz and by using a custom set of electrodes placed on the scalp according to the 10/10 international system, i.e., the electrodes \sloppy{AF$\{3,4,7,8,z\}$, C$\{1,2,3,4,5,6\}$, CP$\{1,2,3,4,5,6,z\}$, Cz, F$\{1,2,3,4,5,6,7,8\}$, Fp$\{1,2,z\}$, FC$\{1,2,3,4,5,6,z\}$, FT$\{7,8\}$, Fz, Iz, O$\{1,2,z\}$, P$\{1,2,3,4,5,6,7,8\}$, PO$\{3,4,7,8,z\}$, Pz, T$\{7,8,9,10\}$, TP$\{7,8\}$}, which are depicted in Fig. \ref{fig:10-10sys-sel}.\\
	For more details on the experimental paradigm and recording setting, please refer to the dataset B repository website (\url{https://physionet.org/content/eegmmidb/1.0.0/}).\\
	Table \ref{tab:dataset_summary} provides a summary of the two datasets.\\
	\begin{table}[htb!]
		\centering
		\caption{Dataset A and dataset B summary.}
		\begin{tabular}{p{100pt}p{100pt}p{100pt}}
			\hline
			\textbf{Characteristic} & \textbf{dataset A} & \textbf{dataset B}  \\
			\hline
			resource & \url{https://physionet.org/content/eegmat/1.0.0/} & \url{https://physionet.org/content/eegmmidb/1.0.0/} \\
			experiment & cognitive workload & motor/imagery \\
			\# participants & 36 & 109 \\
			\# recordings per subject & 2 & 14 \\
			conditions & REST, MAT & CLOSE, LH, RH \\
			electrode positioning & 10/20 international system & modified 10/10 international system\\
			recording sampling rate & 500 Hz & 160 Hz \\			
			pre-processing & bandpass (0.5 - 45 Hz) and notch (50 Hz) filtering & none \\
			\hline
		\end{tabular}
		\label{tab:dataset_summary}
	\end{table}
	%
	
	% EEG FEATURE COMPUTATION
	\subsection{Feature computation}\label{sec:eeg_feature_computation}
	Wanting to provide a heterogeneous set of features in different domains, we consider the following features: the Hjorth parameters \citep{hjorth1970eeg, rodriguez2013efficient}, the power spectral density (PSD) estimated through Welch's method \citep{welch1967use} and the PSD extracted through Morlet wavelet \citep{cohen2014analyzing, cohen2019better}, detailed in the following and developed in MATLAB.\\
	The first feature type is represented by the Hjorth activity, mobility and complexity parameters \citep{hjorth1970eeg, rodriguez2013efficient}, which provide a characterization of the EEG signal in the time domain, while containing information about the signal frequency spectrum. In fact, following the formulae described by \cite{oh2014novel}:
	\begin{itemize}
		\item The activity parameter corresponds to the signal power representation and is computed as the variance of the EEG signal $x(t)$
		\begin{equation}\label{formula:activity}
		activity(x(t)) = var(x(t));
		\end{equation}
		\item The mobility parameter corresponds to the mean frequency and is the square root of the variance of the first derivative of the signal divided by the activity parameter
		\begin{equation}\label{formula:mobility}
		mobility(x(t)) = \sqrt{\frac{activity(x'(t))}{activity(x(t))}};
		\end{equation}
		\item the complexity parameter corresponds to the change of frequency representation and thus is the ratio between the mobility computed on the first derivative of the signal and the mobility computed on the original signal
		\begin{equation}\label{formula:complexity}
		complexity(x(t)) = \frac{mobility(x'(t))}{mobility(x(t))}.
		\end{equation}
	\end{itemize}
	Moving from the time to the frequency domain, we provide a PSD estimation through Welch's method \citep{welch1967use} on the frequency bands of interest characterizing the EEG signal. This technique estimates the PSD by \citep{stoica2005spectral}:
	\begin{enumerate}
		\item Dividing the $x(t)$ characterized by $N$ samples into $S$ segments of $M$ samples each, where 
		\begin{equation*}
		x_j(t) = x((j - 1)K + t)
		\end{equation*}
		is the $j^{th}$ data segment of $x(t)$, having $t = 1, ..., M$, $j = 1, ..., S$, $K$ determines the segments overlapping and $(j - 1)K$ is the $j^{th}$ segment starting point;
		\item Computing the windowed periodogram of each segment, where 
		\begin{equation*}
		\phi_j(f) = \frac{1}{MP} \left|\sum_{t=1}^{M} w(t)x_j(t)e^{-i f t} \right|^2
		\end{equation*}
		is the windowed periodogram of $x_j(t)$, $f$ is the frequency band on which the periodogram is computed, $P = 1/M \sum_{t=1}^{M} |w(t)|^2$ is the power of $w(t)$ and $w(t)$ is the temporal window.
		\item Averaging the obtained windowed periodograms
		\begin{equation}\label{formula:psd_welch}
		PSD_{Welch}(f) = \frac{1}{S} \sum_{j=1}^{S} \phi_j(f).
		\end{equation}
	\end{enumerate}
	In our study we estimate the PSD through Welch's method by applying a 50\% overlap ($K = M/2$) of subsequent segments and by using a Hamming window \citep{stoica2005spectral}.\\
	Moreover, we are interested in the localization of the frequency information in time, thus we use the Morlet wavelet, following the definition given by \cite{cohen2014analyzing, cohen2019better}.\\ 
	A complex Morlet wavelet is defined as 
	\begin{equation*}
	cmw(t) = e^{2 i \pi f t}e^{\frac{-t^2}{2 \sigma^2}},
	\end{equation*}
	where $i = \sqrt{-1}$, $f$ is the frequency (Hz), $t$ is the time (s) and the complex Morlet wavelet is centered at $t = 0$ to avoid phase shift, $\sigma = n/(2 \pi f)$ is the width of the Gaussian window. $n$ contains the information about the number of cycles, which controls the time-frequency precision trade-off. 
	Low values of $n$ correspond to a better time precision, while high values of $n$ correspond to a better frequency precision. \\ 
	In our previous study \cite{saibene2020centric}, we have considered different values for the $n$ parameter in order to explore the differences between the Morlet features having \citep{cohen2014analyzing} a better trade-off between time and frequency precision.
	We have verified that the Morlet wavelet with the trade-off between the time and frequency precision was selected more times in respect to the other Morlet features. Therefore, in this work, we use $n = [3 - 7]$.\\
	After having computed the complex Morlet wavelet, the Fast Fourier Transform (FFT) \citep{frigo1998fftw} is applied to the $cmw(t)$ and its resulting amplitude normalized in the frequency domain 
	\begin{equation*}
	cmwX(f) = \frac{FFT(cmw)}{max(FFT(cmw))}.
	\end{equation*}
	This ensures that the following convolution result is in the same unit of the original EEG signal
	\begin{equation*}
	cx(f, t) = iFFT(FFT(x(t)) .* cmwX),
	\end{equation*}
	that produces the 2 dimensional time-frequency signal and where iFFT corresponds to the inverse FFT.\\
	The power spectral density is then obtained by integrating the power data of the frequency band $\Delta f$
	\begin{equation}\label{formula:psd_morlet}
	PSD_{Morlet}(t) = \sum_{\Delta f} |cx(f, t)|^{2}.
	\end{equation}
	Therefore, the described features are processed for each channel and each feature in the frequency domain is evaluated on the frequency bands of interest, obtaining the final data matrix whose rows and columns correspond to the data instances and the features, respectively.\\
	
	% EEG FEATURE SELECTION
	\section{Genetic algorithm feature selection}\label{sec:eeg_feature_selection}
\begin{figure}[htbp!]
		\centering
		\includegraphics[width=0.6\textwidth]{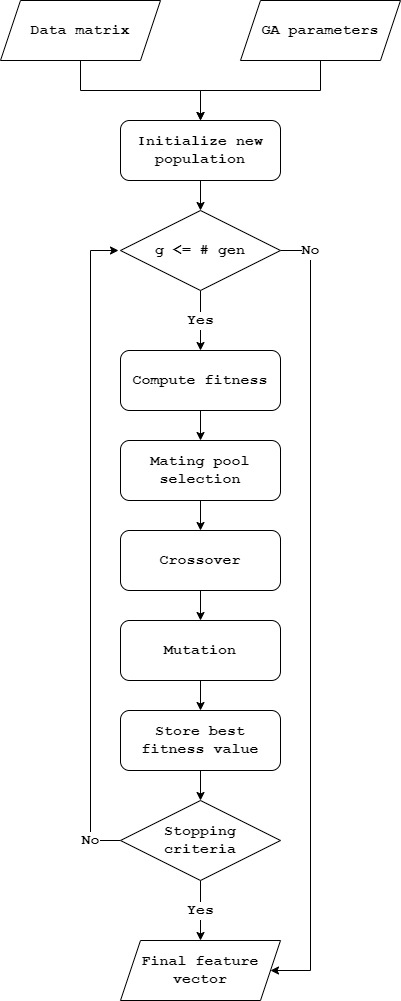}
		\caption{Flowchart of the developed genetic algorithm.}
		\label{fig:GA_feature_selection_fc}
	\end{figure}

	The EEG feature selection is performed through the application of a GA, whose flowchart is depicted in Fig. \ref{fig:GA_feature_selection_fc}.
	
	\noindent{Notice that this process takes as inputs (i) the data matrix represented by the instances of each task of interest performed by the analyzed subjects and by the features extracted from them, and (ii) the classical GA parameters, i.e., the population solution size $n_p = 8$, the number of parents in the mating pool $p_m = 4$ and the number of mutations $n_m = 3$ \citep{saibene2020centric}.}\\ 
	The generated output, or final population, is a binary vector, where 1s correspond to the selected features, 0s otherwise.\\
	Our GA for feature selection is a MATLAB modified implementation of the GA described by \cite{mitchell1998introduction}:
	\begin{enumerate}
		\item The initial population is randomly initialized as a matrix of dimension $n_c \times N_{if}$, where $n_c$ corresponds to the number of chromosomes (i.e., candidate solutions to the feature selection problem), and  $N_{if}$ to the initial dimension of the feature vector;
		\item For each chromosome $c$, the fitness $f(c)$ is computed. The fitness function changes accordingly to the experiment we want to perform;
		\item The mating pool is selected by searching the best fitness obtained on the initial population, thus the parents for the new population are generated;
		\item The crossover generates an offspring from the obtained parents, by taking half of the chromosome from a parent and the remaining half from the other one, thus with a uniform crossover \citep{amarasinghe2016eeg};
		\item Each of the offspring chromosomes has a random mutation of $n_m$ genes, i.e., if an element of the chromosome is set to 0 it becomes 1 and vice versa;
		\item The new population is represented by the obtained parents and their offspring;
		\item Steps 2-6 are repeated until the maximum number of generations or one of the custom stopping criteria is met.
	\end{enumerate} 
	Therefore, we maintained the main computation steps, but we developed custom fitness functions and stopping criteria, following described in detail.
	
	\subsection{Fitness function computation}\label{sec:fitness_functions}
\begin{figure}[htb!]
		\centering
		\includegraphics[width=0.8\textwidth]{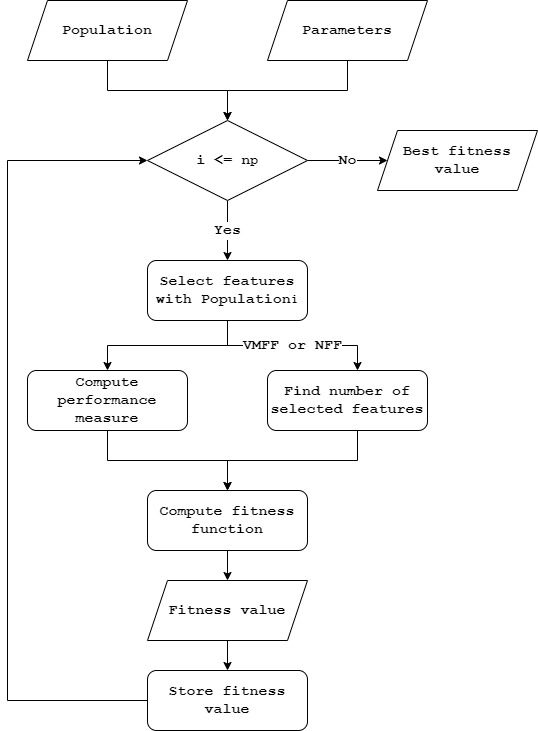}
		\caption{Flowchart of the fitness computation procedure.}
		\label{fig:fitness-procedure}
	\end{figure}

	Fig. \ref{fig:fitness-procedure} depicts the fitness computation procedure, whose inputs are the population, i.e., a $n_p \times N_{if}$ matrix whose rows represent a binary solution for the feature selection, and some parameters that correspond to the learning model and fitness function of interest.\\ 
			
	\noindent{The procedure iterates on the population rows and the features corresponding to each solution are selected from the starting data matrix. Afterwards, the reduced data matrix is passed as input to one of the considered learning models, i.e., a SVM \citep{hastie2009elements} with linear kernel using a 10-fold cross validation as the supervised model, and a K-means clustering algorithm \citep{arthur2006k} with City Block distance \citep{bora2014effect} as the unsupervised one. Then the fitness function of interest is computed using a wrapper \citep{kohavi1998wrapper} approach, obtaining a fitness value for each solution. The best one is stored for each GA generation.}\\
	Describing in more details the fitness function of interests, the first ones correspond to (i) the accuracy \citep{handelman2019peering}, which is an external evaluation measure for the linear SVM
	\begin{equation}\label{formula:accuracy}
	f_{S1}(c) = accuracy = \frac{\text{number of correctly classified instances}}{\text{total number of intances}},
	\end{equation}
	and	(ii) the average silhouette \citep{rousseeuw1987silhouettes}, which is an internal evaluation measure for the K-means clustering algorithm 
	\begin{equation}\label{formula:silhouette}
	f_{U1}(c) = s(e) = \frac{b(e) - a(e)}{max\{a(e), b(e)\}}, -1 \leq s(e) \leq 1,
	\end{equation}
	where it is assumed that there is more than one cluster, $s(e)$ is the silhouette of the cluster element $e$, $a(e)$ represents the average dissimilarity that the element $e$, assigned to cluster $A$, has in respect to the other elements in $A$, $b(e)$ is the minimum mean distance that $e$ has in respect to the other elements present in the other clusters. From now on, when referring to Eq. \ref{formula:accuracy} and \ref{formula:silhouette}, we use the acronym POFF (Performance Only Fitness Function). \\ 
	However, finding only the best performance obtained by a learning model may not represent the best solution in terms of feature vector minimization. Wanting to select the features for mortality prediction of septic patients, \cite{vieira2013modified} proposed a fitness function for their modified PSO that considers both a SVM classifier performance ($P$) and the ratio between the number of selected features ($N_{sf}$) and the initial number of features ($N_{if}$):
	\begin{equation}\label{formula:vieira}
		f(c) = \lambda (1 - P) + (1 - \lambda) \left(1 - \frac{N_{sf}}{N_{if}} \right),
	\end{equation}
	where $\lambda \in [0, 1]$ is a constant parameter that weights the performance-feature number trade-off.\\
	Our proposed fitness functions to consider the best trade-off between the number of selected features and the measures in Eq. \ref{formula:accuracy} and \ref{formula:silhouette}, are then a modified version of Eq. \ref{formula:vieira}:
	\begin{equation}\label{formula:viera_sup}
	f_{S2}(c) = \lambda (1 - f_{S1}(c)) + (1 - \lambda) \left(1 - \frac{N_{sf}}{N_{if}} \right),
	\end{equation}
	\begin{equation}\label{formula:viera_unsup}
	f_{U2}(c) = \lambda (1 - f_{U1}(c)) + (1 - \lambda) \left(1 - \frac{N_{sf}}{N_{if}} \right),
	\end{equation}
	where Eq. \ref{formula:viera_sup} and \ref{formula:viera_unsup} are performed if the model is supervised or unsupervised, respectively. 0.88 is chosen as the default value for $\lambda$, having verified that other values of $\lambda$ do not provide significant better results as reported in our previous work \citep{saibene2020centric}. Eq. \ref{formula:viera_sup} and \ref{formula:viera_unsup} fitness function type is called VMFF (Vieira Modified Fitness Function).\\
	Finally, our novel fitness function type wants to directly use accuracy and silhouette to weight the number of selected features in respect to the initial feature vector dimension, thus
	\begin{equation}\label{formula:f3_sup}
	f_{S3}(c) = f_{S1}(c) \left( 1 - \frac{N_{sf}}{N_{if}} \right),
	\end{equation}
	\begin{equation}\label{formula:f3_unsup}
	f_{U3}(c) = f_{U1}(c) \left( 1 - \frac{N_{sf}}{N_{if}} \right).
	\end{equation}
	Therefore, these fitness functions greatly penalize large feature vectors and we refer to them as NFF (Novel Fitness Function).\\
	Notice that the Eq. \ref{formula:viera_sup}, \ref{formula:viera_unsup}, \ref{formula:f3_sup} and \ref{formula:f3_unsup} consider the feature selection problem with a multi-objective \citep{konak2006multi} approach, wanting to find the smaller feature vector needed to obtain high values of accuracy or silhouette.\\
	The described fitness functions are summarized in Table \ref{tab:ff_summary}.\\
	\begin{table}[htb!]
		\centering
		\caption{Summary of the employed fitness functions.}
		\resizebox{\textwidth}{!}{
		\begin{tabular}{lll}%p{30pt}p{135pt}p{135pt}
			\hline
			\textbf{Type} & \textbf{Supervised} & \textbf{Unsupervised}  \\
			\hline
			POFF & $f_{S1}(c) = \frac{\text{\# correctly classified instances}}{\text{total \# instances}}$ & $f_{U1}(c) = \frac{b(e) - a(e)}{max\{a(e), b(e)\}}$ \\
			VMFF & $f_{S2}(c) = \lambda (1 - f_{S1}(c)) + (1 - \lambda) \left(1 - \frac{N_{sf}}{N_{if}} \right)$ & $f_{U2}(c) = \lambda (1 - f_{U1}(c)) + (1 - \lambda) \left(1 - \frac{N_{sf}}{N_{if}} \right)$ \\
			NFF & $f_{S3}(c) = f_{S1}(c) \left( 1 - \frac{N_{sf}}{N_{if}} \right)$ & $f_{U3}(c) = f_{U1}(c) \left( 1 - \frac{N_{sf}}{N_{if}} \right)$ \\
			\hline
		\end{tabular}
		}
		\label{tab:ff_summary}
	\end{table}
	
	\noindent{To conclude this section, we highlight that the introduction of a 10-fold cross validation in the supervised approach is purposely made to provide more robustness on the SVM model for each generation. Therefore, a faster computation is sacrificed to obtain more reliable results.}
		
	\subsection{Stopping criteria}\label{sec:stopping_criteria}
	Apart from the classical stopping criterion represented by the maximum number of generations, set as a parameter of the GA, we introduced constraints on the computation time and on the performance achieved after a certain number of iterations.\\
	In fact, a maximum time parameter is added to limit the allowed computation time, considering the possible necessity of analyzing the data between experimental sessions and having limited computational resources.\\
	The inclusion of some checks on the performance, was also introduced to allow a dynamic change of the maximum number of generations (initially set to 200), by tracking the local and global bests of the obtained fitness values:
	\begin{enumerate}
		\item The local best, i.e., the best fitness value obtained during an iteration, substitutes the global best, when it is greater than the global best;
		\item If after the $80\%$ of generations the local best has not substituted the global best, the iterations stop;
		\item We introduce a fitness check variable that is updated after the $50\%$ of the maximum number of generations and corresponds to the best fitness value obtained during the whole computation; if the best fitness value is greater than the fitness check value plus $0.02$, then the maximum number of generations is increased by the $50\%$;
		\item If the maximum number of generations is greater than 1000 and the best fitness value is equal to 1, the iterations stop.
	\end{enumerate}
	The percentages and number of iterations have been set empirically.

	% % % % % RESULTS % % % % %
	\section{Results and discussion}\label{sec:results}
	In this section the results obtained following the procedure depicted in Fig. \ref{fig:our_proposal_fc} are reported and discussed. Therefore, the data matrices obtained by pre-processing, normalizing and computing the features on dataset A, B and on a mix of the two are detailed and the conducted experiments described. 
	
	\subsection{Data matrices generation}
\begin{figure}[htb!]
		\centering
		\includegraphics[width=0.6\textwidth]{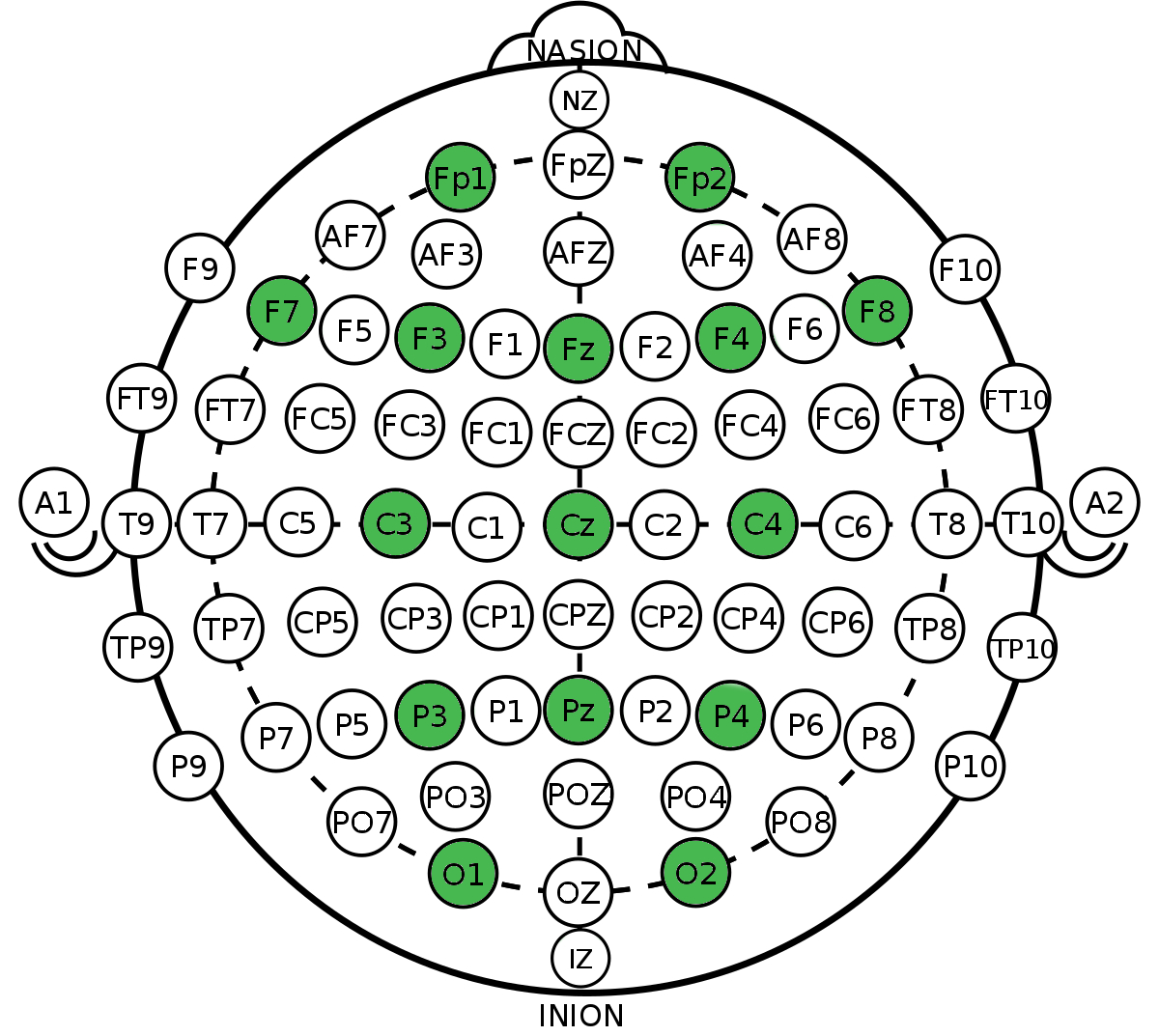}
		\caption{Dataset A and B common electrodes.}
		\label{fig:gen-electrodes}
	\end{figure}

	Starting from the original EEG data of dataset A and B (Section \ref{sec:datasets}), 3 data matrices have been generated: data matrix A, B and GEN.\\
	Notice that all the features described in Section \ref{sec:eeg_feature_computation}, i.e., activity, mobility and complexity parameters, PSD estimated through the Welch's method and PSD computed on the time-frequency signals obtained through the Morlet wavelet convolution, are computed on each dataset.\\
	To obtain the data matrix A, z-score normalization is performed on the original mental workload EEG data. The recordings of subject 4 and 31 were excluded due to the presence of an incorrect number of samples. The recordings of the remaining 34 subjects are analyzed.\\ Afterwards, the features are computed on the 19 channels depicted in Fig. \ref{fig:10-20sys-sel} and considering the following frequency bands: the lower $\theta$ ($\theta_l = [4.1 - 5.8]$ Hz), higher $\theta$ ($\theta_h = [5.9 - 7.4]$ Hz), lower $\beta$ ($\beta_l = [13 - 19.9]$ Hz) and higher $\beta$ ($\beta_h = [20 - 25]$ Hz) rhythms. In fact, we followed the suggestions given by \cite{seleznov2019detrended}, who affirm that $\theta$ and $\beta$ frequency bands bring the cognitive processes involved in the mental calculation, considering the managing of memory, consciousness and the task influence on the subjects’ emotional state.\\
	Therefore, data matrix A presents 68 rows, i.e., 2 instances per analyzed subject that correspond to the REST and MAT tasks, and 209 columns, representing the computed features: $\text{19 electrodes} \times (\text{3 Hjorth parameters} + \text{4 frequency bands} \times (\text{1 } PSD_{Welch} + \text{1 } PSD_{Morlet}))$.\\
	To obtain the data matrix B, the original motor movement/imagery EEG signals are firstly bandpass filtered in the range $[0.5 - 45]$ Hz and notch filtered at 50 Hz. We mimicked the pre-processing performed by the authors of dataset A to have more comparable data. Subsequently, the data are z-score normalized and the features computed on the 64 electrodes depicted in Fig. \ref{fig:10-10sys-sel} and considering the $\theta$, $\alpha$ and $\beta$ frequency bands. The rhythm choice was due to the necessity of keeping track not only of the cognitive workload, but also of the difference between the resting states and the motor movement tasks.\\
	Therefore, data matrix B dimensions are $5033 \times 576$. The number of rows corresponds to all the subjects' recordings of dataset B and their considered tasks, i.e., 109 tasks of CLOSED, 2469 tasks of LH and 2455 tasks of RH. The columns report the computed features: $\text{64 electrodes} \times (\text{3 Hjorth parameters} + \text{3 frequency bands} \times (\text{1 } PSD_{Welch} + \text{1 } PSD_{Morlet}))$.\\
	Finally, we combined dataset A and B to obtain dataset GEN and test the generalizability of our proposal. All the analyzed instances of dataset A are retained, while 34 instances for each condition are selected randomly from dataset B. The selected subjects are 1, 2, 5, 6, 9, 10, 12,	15-20, 22, 25, 27-29, 32, 34, 37-39, 41, 42, 44, 46-53, 56, 58-62, 64, 66, 68, 71-73, 75, 77, 78, 80, 82-85, 87-92, 94, 96, 98, 99-104, 107-109.\\
	Subsequently, the feature computation is performed on the dataset A and B common set of electrodes (Fig. \ref{fig:gen-electrodes}) and considering the $\theta$, $\alpha$ and $\beta$ frequency bands.\\
	 
	\noindent{Therefore, data matrix GEN presents 170 rows, i.e., 34 instances of each task (REST, MAT, CLOSED, LH, RH), and 135 columns, representing the computed features: $\text{15 electrodes} \times (\text{3 Hjorth parameters} + \text{3 frequency bands} \times (\text{1 } PSD_{Welch} + \text{1 } PSD_{Morlet}))$.}

	\subsection{Experiments}
	The 3 data matrices A, B and GEN are inputted to the feature vector reduction strategies depicted in Fig. \ref{fig:our_proposal_fc}: our proposed GA for feature selection is compared to the benchmark, constituted by the application of the learning models on the data matrices retaining all the computed features (ALL) and of the PCA technique.\\
	In this section are reported and discussed the results obtained by applying the supervised and unsupervised approaches on data matrices A, B and GEN, using the fitness functions described in Section \ref{sec:fitness_functions}. Thus, the performed experiments are the ones present in Table \ref{tab:list_experiments}.\\
	\begin{table}[htb!]
		\centering
		\caption{List of experiments.}
		\resizebox{\textwidth}{!}{%
		\begin{tabular}{p{70pt}p{70pt}p{70pt}p{70pt}}
			\toprule
			\textbf{Data matrix} & \textbf{Supervised} & \textbf{Unsupervised} & \textbf{Fitness type} \\
			\midrule
			A & $f_{S1}(c)$ & $f_{U1}(c)$ & POFF \\
			A & $f_{S2}(c)$ & $f_{U2}(c)$ & VMFF \\
			A & $f_{S3}(c)$ & $f_{U3}(c)$ & NFF \\			
			B & $f_{S1}(c)$ & $f_{U1}(c)$ & POFF \\
			B & $f_{S2}(c)$ & $f_{U2}(c)$ & VMFF \\
			B & $f_{S3}(c)$ & $f_{U3}(c)$ & NFF \\
			GEN & $f_{S1}(c)$ & $f_{U1}(c)$ & POFF \\
			GEN & $f_{S2}(c)$ & $f_{U2}(c)$ & VMFF \\
			GEN & $f_{S3}(c)$ & $f_{U3}(c)$ & NFF \\
			\bottomrule
		\end{tabular}
		}
		\label{tab:list_experiments}
	\end{table}
	
	\noindent{Moreover, the computation time and generation number for each supervised and unsupervised experiment are analyzed to understand the efficiency and efficacy of the proposed methods. }

	\subsection{GA feature selection based on the supervised approach}
	\label{sec:supervised_results}
	The fitness values obtained by applying our GA for feature selection with a supervised approach are reported in Table \ref{tab:supervised_fv}.\\
	Specifically, in Table \ref{tab:supervised_fv} are listed the fitness values obtained by applying the different fitness functions, with the following meaning:
	\begin{itemize}
		\item \textit{mean} is the mean of the fitness values obtained by each generation;
		\item \textit{std} is the standard deviation of the fitness values obtained by each generation;
		\item \textit{max} is the maximum of the fitness values obtained by each generation;
		\item \textit{final} is the maximum value present in the range $[mean - std, mean + std]$ of the fitness values obtained by each generation. This value is considered as the most representative of our GA for feature selection.
	\end{itemize}	
	Finally, fields \textit{GAFS} and \textit{Benchmark} are used to discriminate the results obtained by using our GA for Feature Selection and the benchmarking techniques, respectively.\\
	%
	% SUPERVISED - FITNESS
	\begin{table}[htb!]
		\centering
		\caption{Fitness values obtained on data matrices A, B and GEN.}
		\resizebox{\textwidth}{!}{%
			\begin{tabular}{p{60pt}p{40pt}p{40pt}p{40pt}p{50pt}p{40pt}p{40pt}}
				\toprule
				& \multicolumn{4}{l}{\textbf{GAFS}} & \multicolumn{2}{l}{\textbf{Benchmark}} \\
				\midrule
				\textbf{ID} & \textbf{mean} & \textbf{std} & \textbf{max} & \textbf{final} & \textbf{ALL} & \textbf{PCA} \\
				\hline
				A-$f_{S1}(c)$  & 1.00 & 0.0000 & 1.00 & 1.00 & 1.00 & 1.00 \\
				A-$f_{S2}(c)$  & 0.09 & 0.0131 & 0.12 & 0.10 & 0.00 & 0.12 \\
				A-$f_{S3}(c)$  & 0.78 & 0.0688 & 0.86 & 0.85 & - 	& 0.95 \\
				\hline
				B-$f_{S1}(c)$  & 0.90 & 0.0021 & 0.90  & 0.90 & 0.90  & 0.57 \\
				B-$f_{S2}(c)$  & 0.18 & 0.0010 & 0.18  & 0.18 & 0.08  & 0.49 \\
				B-$f_{S3}(c)$  & 0.48 & 0.0028 & 0.48  & 0.48 & -	  & 0.56 \\
				\hline
				GEN-$f_{S1}(c)$ & 0.85 & 0.0123 & 0.88 & 0.86 & 0.81 & 0.66 \\
				GEN-$f_{S2}(c)$ & 0.29 & 0.0252 & 0.34 & 0.32 & 0.16 & 0.38 \\
				GEN-$f_{S3}(c)$ & 0.69 & 0.0781 & 0.80 & 0.77 & -	 & 0.64 \\
				\bottomrule
			\end{tabular}%
		}
		\label{tab:supervised_fv}%
	\end{table}%
	
	\noindent{The fitness values are analyzed to better understand the evolution of the selection process performed by our proposal and they may vary depending on the employed fitness function. Reporting the performance measures is then necessary to evaluate the goodness of the SVM model on the data matrix subsets. Therefore, small fitness values do not necessarily correspond to bad performances.} \\
	
	In Table \ref{tab:supervised_fv}, notice that the fitness values corresponding to column \textit{ALL} and fitness function $f_{S3}(c)$ are not reported. In fact, this fitness function (Eq. \ref{formula:f3_sup}) is designed to avoid large feature vectors and returns 0 when all the features are selected. Thus, it is meaningless to perform this test in the \textit{ALL} case.\\
	The experiment A-$f_{S1}(c)$ returns the maximum fitness value for each column, thus it can be assumed that data matrix A presents easily classifiable conditions, being this fitness function $f_{S1}(c)$ (Eq. \ref{formula:accuracy}) equal to the accuracy of the SVM model.\\
	Moreover, B-$f_{S1}(c)$ and GEN-$f_{S1}(c)$ obtains generally high fitness values in each column, except for the PCA benchmarking technique. Therefore, our proposal seems to provide a better representation of the data in respect to PCA.\\
	The $f_{S2}(c)$ introduces a weighting of the trade-off between the accuracy and number of selected features through the $\lambda = 0.88$ parameter. It generally returns lower fitness values compared to the other fitness functions, however we remind that it does not automatically mean that the SVM model achieves poor results on the performance measures. \\
	Finally, the fitness values obtained by experiments $f_{S3}(c)$ have a greater variability (column \textit{std}) in respect to the other fitness functions, that may be due to the use of the accuracy as a weighting factor for the $N_{sf}$ and $N_{if}$ ratio.	\\
	
	Wanting to understand the effectiveness of the \textit{final} solution feature subset, Table \ref{tab:supervised_A} to \ref{tab:supervised_GEN} report the number of selected features (\textit{$N_{sf}$}), the accuracy for each data matrix task (\textit{Acc}), the global accuracy (\textit{gAcc}) and the weighted-average F1-score (\textit{waF1}), which corresponds to the mean of the F1-scores of each class weighted with each class number of instances.\\
	Recall that the classification problem (i) is binary for data matrix A, (ii) considers 3 classes for data matrix B and (iii) 5 for data matrix GEN.\\
	%
	% OTHER MEASURES
	% EDMAT - SUPERVISED - TAB 2
	\begin{table}[htb!]
		\caption{Classification performances on data matrix A.}
		\resizebox{\textwidth}{!}{%
			\begin{tabular}{p{60pt}p{50pt}p{40pt}p{50pt}p{40pt}p{40pt}}
				\toprule
				\textbf{ID} & $\boldsymbol{N_{sf}}$ & \multicolumn{2}{l}{\textbf{Acc}} & \textbf{gAcc} & \textbf{waF1}  \\
				\cline{3-4}
				& & \textbf{MAT} & \textbf{REST} & & \\
				\hline
				A-ALL			& 209 & 1.00 & 1.00 & \textbf{1.00} & \textbf{1.00} \\
				A-PCA			&	8 & 1.00 & 1.00 & \textbf{1.00} & \textbf{1.00} \\
				A-$f_{S1}(c)$ 	& 106 & 1.00 & 1.00 & \textbf{1.00} & \textbf{1.00} \\
				A-$f_{S2}(c)$	&  35 & 1.00 & 1.00 & \textbf{1.00} & \textbf{1.00} \\
				A-$f_{S3}(c)$	&  32 & 1.00 & 1.00 & \textbf{1.00} & \textbf{1.00} \\
				\bottomrule 
			\end{tabular}
		}
		\label{tab:supervised_A}%
	\end{table}%
	\begin{table}[htb!]
		\caption{Classification performances on data matrix B.}
		\resizebox{\textwidth}{!}{%
			\begin{tabular}{p{60pt}p{50pt}p{40pt}p{40pt}p{50pt}p{40pt}p{40pt}}
				\toprule
				\textbf{ID} & $\boldsymbol{N_{sf}}$ & \multicolumn{3}{l}{\textbf{Acc}} & \textbf{gAcc} & \textbf{waF1}  \\
				\cline{3-5}
				& & \textbf{CLOSE} & \textbf{LH} & \textbf{RH} & & \\
				\hline
				B-ALL			& 576 & 1.00 & 0.90 & 0.90 & \textbf{0.93} & \textbf{0.90} \\
				B-PCA			&  13 & 1.00 & 0.57 & 0.57 & 0.71 & 0.52 \\
				B-$f_{S1}(c)$ 	& 293 & 1.00 & 0.89 & 0.89 & 0.89 & 0.89 \\
				B-$f_{S2}(c)$	& 291 & 1.00 & 0.87 & 0.87 & 0.87 & 0.87 \\
				B-$f_{S3}(c)$	& 256 & 1.00 & 0.89 & 0.89 & 0.89 & 0.88 \\
				\bottomrule
			\end{tabular}
		}
		\label{tab:supervised_B}%
	\end{table}%
	\begin{table}[htb!]
		\caption{Classification performances on data matrix GEN.}
		\resizebox{\textwidth}{!}{%
			\begin{tabular}{p{60pt}p{50pt}p{40pt}p{40pt}p{40pt}p{40pt}p{50pt}p{40pt}p{40pt}}
				\toprule
				\textbf{ID} & $\boldsymbol{N_{sf}}$ & \multicolumn{5}{l}{\textbf{Acc}} & \textbf{gAcc} & \textbf{waF1}  \\
				\cline{3-7}
				& & \textbf{MAT} & \textbf{REST} & \textbf{CLOSE} & \textbf{RH} & \textbf{LH} & & \\
				\hline
				GEN-ALL			& 135 & 1.00 & 1.00 & 1.00 & 0.81 & 0.81 & 0.82 & 0.81 \\
				GEN-PCA			&   5 & 0.89 & 0.98 & 0.90 & 0.78 & 0.78 & 0.66 & 0.64 \\
				GEN-$f_{S1}(c)$ &  77 & 1.00 & 1.00 & 1.00 & 0.86 &	0.86 & 0.86 & 0.86 \\
				GEN-$f_{S2}(c)$	&  51 & 1.00 & 1.00 & 1.00 & 0.75 &	0.75 & 0.75 & 0.75 \\
				GEN-$f_{S3}(c)$	&  19 & 1.00 & 1.00 & 1.00 & 0.92 & 0.92 & \textbf{0.92} & \textbf{0.92} \\
				\bottomrule 
			\end{tabular}
		}
		\label{tab:supervised_GEN}%
	\end{table}%

	\noindent{The classification performances on data matrix A are coherent with the results reported in Table \ref{tab:supervised_fv} for the $f_{S1}(c)$ fitness function, where all the performance measures achieve the maximum score. We highlight that our approach presents a great decrease in the selected feature number, especially for the VMFF and NFF experiments, thus avoiding the possibility of incurring in the overfitting problem, having a data matrix of dimension $68 \times 209$.}\\
	For what concerns data matrix B, our proposal presents a significant decrease in the number of features (around the $50\%$ of the initial number of features) and performance measures comparable with the ones obtained considering all the initial features.\\
	Notice that the CLOSE task is always detected, while the LH and RH are sometimes confused. This missclassification may be due to the difficulty of the movement tasks, the presence of outliers and the incorrect performance of movement. In fact, there are no guarantees that a subject performed correctly the required task.\\ 
	Finally, the results obtained on the GEN data matrix are in line with the ones reported for data matrices A and B. The MAT, REST and CLOSE tasks are always recognized, while there are some misclassifications for the RH and LH tasks. Even though there are 5 classes of which 2 comes from dataset A and 3 from dataset B, and the classification task is more complex, the performance remain high. Moreover, notice that the best global performance is achieved by our proposal in the NFF case, which uses about the $14\%$ of the original features to obtain a global accuracy of $0.92$, thus increasing the global accuracy achieved on the GEN-ALL case by $0.10$.\\
	
	As a final analysis on the results, the computation time and the generation number for each experiment are listed in Table \ref{tab:time_supervised}.
	\begin{table}[htb!]
		\centering
		\caption{Computation time and generation number for each experiment.}
		%\resizebox{\textwidth}{!}{%
		\begin{tabular}{p{60pt}p{60pt}p{60pt}p{60pt}}
			\toprule
			\textbf{Machine} & \textbf{Test} & \textbf{Minutes} & \textbf{Generations}\\
			\midrule
			\multirow{3}{*}{2} &
			A-$f_{S1}(c)$ & 28.1691 & 200 \\
			& A-$f_{S2}(c)$ & 14.4717 & 240 \\
			& A-$f_{S3}(c)$ & 17.6436 & 300 \\
			\hline
			\multirow{3}{*}{1} &
			B-$f_{S1}(c)$ & 670.5478 & 7 \\
			& B-$f_{S2}(c)$ & 644.4401 & 6 \\
			& B-$f_{S3}(c)$ & 686.5514 & 7 \\
			\hline
			\multirow{3}{*}{2} &
			GEN-$f_{S1}(c)$ & 51.3699 & 240 \\
			& GEN-$f_{S2}(c)$ & 60.1090 & 313 \\
			& GEN-$f_{S3}(c)$ & 60.1604 & 302 \\
			\bottomrule
		\end{tabular}%
		%}
		\label{tab:time_supervised}%
	\end{table}%
	
	\noindent{The GA for data matrix B feature selection was performed on an \textit{Intel(R) Core(TM) i7-4790 CPU \@ 3.60 GHz with 16.00 GB RAM}, called \textit{Machine 1}, for the other data matrices on an \textit{Intel(R) Core(TM) i5-4210U CPU \@ 1.70 2.40 GHz with 8.00 GB RAM}, called \textit{Machine 2}.}\\
	As stated in Section \ref{sec:intro}, the use of evolutionary algorithms for feature selection is computationally demanding, especially when designed with a wrapper approach and dealing with high dimensional data.\\
	In fact, for data matrix B ($5033 \times 576$ dimensions) the maximum time parameter is set to 600 minutes and is the criterion on which all the experiments stop. In fact, exploring the other stopping criteria (Section \ref{sec:stopping_criteria}) the final number of performed generation is 7 and most probably the conditions for the fitness value checks are never met. Even though the maximum fitness value is never met, the performance measures reported in Table \ref{tab:supervised_B} are high. \\
	Instead, the experiments on data matrix A ($68 \times 209$) and GEN ($170 \times 135$) present a greater number of generations, even though the imposed value for the maximum time parameter is equal to 60 minutes. For data matrix A the triggered stopping criterion is on the fitness value check, while for data matrix GEN the maximum computation time is exceeded. Nevertheless, the performance obtained by these experiments are comparable to or higher then the ones achieved by the benchmarking techniques, thus the time constraint seems to be suitable for the GA feature computation on data with the reported dimensionalities.\\
	The computational complexity of the GA for feature selection is again confirmed, noticing that with lower computation time and resources the experiments on data matrices A and GEN increased the initial number of generation (200) in respect to the ones performed on data matrix B. 
	
	\subsection{GA feature selection based on the unsupervised approach}
	\label{sec:unsupervised_results}
	The fitness values obtained by applying our GA for feature selection with an unsupervised approach are reported in Table \ref{tab:unsupervised_fv}.\\
	The table fields follow the same criteria explained at the beginning of Section \ref{sec:supervised_results}.\\
	%
	% EDMAT - UNSUPERVISED - TAB 1
	\begin{table}[htb!]
		\centering
		\caption{Fitness values obtained on data matrices A, B and GEN.}
		\resizebox{\textwidth}{!}{%
			\begin{tabular}{p{60pt}p{40pt}p{40pt}p{40pt}p{50pt}p{40pt}p{40pt}}
				\toprule
				& \multicolumn{4}{l}{\textbf{GAFS}} & \multicolumn{2}{l}{\textbf{Benchmark}} \\
				\midrule
				\textbf{ID} & \textbf{mean} & \textbf{std} & \textbf{max} & \textbf{final} & \textbf{ALL} & \textbf{PCA} \\
				\hline
				A-$f_{U1}(c)$   & 0.61  & 0.0227 & 0.65  & 0.63 & 0.54  & 0.43 \\
				A-$f_{U2}(c)$   & 0.54  & 0.0382 & 0.61  & 0.58 & 0.40  & 0.61 \\
				A-$f_{U3}(c)$   & 0.54  & 0.0731 & 0.62  & 0.61 & -  	& 0.42 \\
				\hline
				B-$f_{U1}(c)$   & 0.42  & 0.0295 & 0.46  & 0.45 & 0.31  & 0.36 \\
				B-$f_{U2}(c)$   & 0.85  & 0.0618 & 0.88  & 0.88 & 0.60  & 0.68\\
				B-$f_{U3}(c)$   & 0.33  & 0.0733 & 0.43  & 0.40 & -		& 0.35 \\
				\hline				
				GEN-$f_{U1}(c)$   & 0.60  & 0.0062 & 0.60  & 0.60 & 0.51  & 0.50 \\
				GEN-$f_{U2}(c)$   & 0.71  & 0.0211 & 0.73  & 0.73 & 0.43  & 0.56 \\
				GEN-$f_{U3}(c)$   & 0.52  & 0.0491 & 0.55  & 0.55 & -	  & 0.48 \\
				\bottomrule
			\end{tabular}%
		}
		\label{tab:unsupervised_fv}%
	\end{table}%
	
	\noindent{Notice that we do not make a comparison between the fitness values reported in Table \ref{tab:supervised_fv} and \ref{tab:unsupervised_fv}. In fact, they are the results of a supervised and unsupervised approach, respectively, and also their ranges differ. This is due to the differences between the accuracy and silhouette measures, which varies in the $[0, 1]$ range and $[-1, 1]$ range, respectively.}\\
	Finally, consider that the fitness values are again analyzed to have a better understanding of our GA selection process and their variations are due to the different nature of the employed fitness functions. The performance measures are subsequently reported to evaluate the goodness of the reduced data matrix using a K-means clustering model. Therefore, small fitness values do not necessarily represent poor performances. \\
	
	As in the supervised case, the fitness value of the \textit{ALL} column corresponding to the experiment based on NFF is not reported. Remind that our novel fitness function $f_{U1}(c)$ (Eq. \ref{formula:silhouette}) penalizes the presence of a great number of features in respect to the initial feature vector and thus, when all the features are retained, the fitness value is equal to 0.\\
	A general observation can be done on the results obtained on all the experiments using the $f_{U1}(c)$ fitness function. Being of type POFF, the fitness values correspond to the silhouette scores computed on the clusters generated by the K-means clustering algorithm. An improvement of the silhouette score can be observed on the experiments performed with our GA for feature selection in respect to the one achieved by the benchmarking techniques. \\
	The fitness values obtained by our GA for feature selection are always higher then the one presented in the benchmarking columns (\textit{ALL}, \textit{PCA}) for the unsupervised approach.\\
	
	Wanting to have a better understanding of the results achieved by the final solutions, in Table \ref{tab:unsupervised_A} to \ref{tab:unsupervised_GEN} are reported, for each data matrix and experiment, the silhouette scores (\textit{Silhouette}) of the GA feature selection final solution, and the optimal number of clusters with their corresponding silhouette scores according to a cluster evaluator provided as a native MATLAB function (\textit{evalclusters}).\\
	For the computation of the K-means clustering algorithm in our GA for feature selection, we always set $k$ equal to the number of the dataset tasks. Therefore, $k = 2$ for data matrix A, $k = 3$ for data matrix B and $k = 5$ for data matrix GEN. However, we would like to evaluate if the suggested number of clusters is the optimal in respect to the silhouette score.
	Under \textit{optimal clusters} in Table \ref{tab:unsupervised_A} to \ref{tab:unsupervised_GEN}, the optimal number of clusters according to \textit{evalclusters} are reported. The considered values of $k$ span from 2 to the  number of data matrix tasks plus 1 and the corresponding silhouette scores are reported in a dedicated column. \\
	%
	% Clustering performance A
	\begin{table}[htb!] 
		\centering
		\caption{Clustering performance on data matrix A.}
		\resizebox{\textwidth}{!}{%
			\begin{tabular}{p{60pt}p{30pt}p{60pt}p{40pt}p{70pt}}
				\toprule
				\textbf{ID} &  $\boldsymbol{N_{sf}}$ & \textbf{Silhouette} & \multicolumn{2}{l}{\textbf{Evaluator}} \\
				\cline{3-5}
				&   & \textbf{GAFS} & \textbf{optimal clusters} & \textbf{$\boldsymbol{k=[2 - 3]}$} \\
				\midrule
				A-ALL 		  & 209   & 0.54 & 2     & [0.54 0.46] \\
				A-PCA 		  & 8     & 0.43 & 2     & [0.43 0.41] \\
				A-$f_{U1}(c)$ & 68    & 0.63 & 2     & [0.63 0.49] \\
				A-$f_{U2}(c)$ & 55    & 0.45 & 2     & [0.45 0.38] \\
				A-$f_{U3}(c)$ & 24    & \textbf{0.69} & 2     & [0.69 0.55] \\
				\bottomrule
			\end{tabular}%
		}
		\label{tab:unsupervised_A}%
	\end{table}%
	%
	% Clustering performance B
	\begin{table}[htb!]
		\centering
		\caption{Clustering performance on data matrix B.}
		\resizebox{\textwidth}{!}{%
			\begin{tabular}{p{60pt}p{30pt}p{60pt}p{40pt}p{70pt}}
				\toprule
				\textbf{ID} &  $\boldsymbol{N_{sf}}$ & \textbf{Silhouette} & \multicolumn{2}{l}{\textbf{Evaluator}} \\
				\cline{3-5}
				&   & \textbf{GAFS} & \textbf{optimal clusters} & \textbf{$\boldsymbol{k=[2 - 4]}$} \\
				\midrule
				B-ALL 		  & 576   & 0.31 & 2     & [0.94 0.31 0.11] \\
				B-PCA 		  & 13    & 0.36 & 3     & [0.41 0.43 0.26] \\
				B-$f_{U1}(c)$ & 214   & 0.45 & 2     & [0.92 0.45 0.36] \\
				B-$f_{U2}(c)$ & 199   & 0.08 & 2     & [0.95 0.28 0.28] \\
				B-$f_{U3}(c)$ & 148   & \textbf{0.54} & 2     & [0.90 0.54 0.43] \\
				\bottomrule
			\end{tabular}%
		}
		\label{tab:unsupervised_B}%
	\end{table}%
	%
	% Clustering performance GEN
	\begin{table}[htb!]
		\centering
		\caption{Clustering performance on data matrix GEN.}
		\resizebox{\textwidth}{!}{%
			\begin{tabular}{p{60pt}p{30pt}p{60pt}p{40pt}p{110pt}}
				\toprule
				\textbf{ID} &  $\boldsymbol{N_{sf}}$ & \textbf{Silhouette} & \multicolumn{2}{l}{\textbf{Evaluator}} \\
				\cline{3-5}
				&   & \textbf{GAFS} & \textbf{optimal clusters} & \textbf{$\boldsymbol{k=[2 - 6]}$} \\
				\midrule
				GEN-ALL & 135   & 0.51 & 2     & [0.73 0.58 0.55 0.52 0.52] \\
				GEN-PCA & 5     & 0.50 & 2     & [0.69 0.47 0.48 0.49 0.50] \\
				GEN-$f_{U1}(c)$ & 32    & 0.60 & 2     & [0.75 0.60 0.58 0.59 0.54] \\
				GEN-$f_{U2}(c)$ & 44    & 0.26 & 2     & [0.73 0.57 0.55 0.48 0.48] \\
				GEN-$f_{U3}(c)$ & 14    & \textbf{0.61} & 2     & [0.75 0.62 0.58 0.56 0.53] \\
				\bottomrule
			\end{tabular}%
		}
		\label{tab:unsupervised_GEN}%
	\end{table}%

	\noindent{According to Table \ref{tab:unsupervised_A}, the optimal number of clusters for data matrix A is always 2, which corresponds to the number of its tasks (REST, MAT). A significant improvement of the silhouette score is achieved by the NFF experiment, whose silhouette is $0.15$ points greater then the A-ALL performance measure and whose number of selected features correspond to about the $11\%$ of the original feature vector.}\\
	Instead, the final solutions on data matrix B (Table \ref{tab:unsupervised_B}) present generally lower values of silhouette and the evaluator suggest as the optimal number of clusters $k = 2$. In fact, the performances increase in correspondence of this value. However, our approaches again achieve a general improvement of the silhouette score for $k = 3$ (CLOSE, LH, RH) in respect to the benchmarking techniques. Also, they obtain a significant decrease of the feature vector, about the $26\%$ of the initial number of features for the NFF case.\\
	The unsupervised results are actually in-line with the observations given for the supervised approach. In fact, there was a good number of misclassifications of the LH and RH tasks, which do not seem to be easy to cluster in the K-means clustering case.\\
	For data matrix GEN experiments there is also a similar mismatch between the desired number of clusters ($k = 5$) and the optimal cluster number suggested by the evaluator ($k = 2$). Table \ref{tab:unsupervised_GEN} lists the performance measure values computed on the data matrix reduced according to the final solution.\\
	The silhouette scores are generally around $0.50$, because the silhouette is probably high on average for the well identified clusters.\\
	We expect REST, MAT and CLOSE conditions to be easier to discriminate in respect to the LH and RH tasks, that involve the same cortical areas.\\
	Again, for the NFF test the silhouette score has a $0.10$ improvement in respect to the ALL case and only the $18.9\%$ of the features are selected.\\
	As a final remark on the performance measure analysis, notice that we do not consider the number of elements present in the clusters, wanting to test the GA unsupervised feature selection by pretending to be unaware of the actual data matrix distributions of instances.\\
	However, adding a weighting factor to the silhouette computed on each cluster, when the range of instances for the tasks of interest are known, could lead to an overall improvement of the unsupervised fitness functions.\\

	As a final analysis on the results, the computation time and generation number of each experiment are listed in Table \ref{tab:time_unsupervised}.\\
	As for the supervised approach, the computation for data matrix B are performed on \textit{Machine 1}, while the others are performed on \textit{Machine 2}.\\
	%
	% UNSUPERVISED
	\begin{table}[htb!]
		\centering
		\caption{Computation time and generation number for each experiment.}
		%\resizebox{\textwidth}{!}{%
		\begin{tabular}{p{60pt}p{60pt}p{60pt}p{60pt}}
			\toprule
			\textbf{Machine} & \textbf{Test} & \textbf{Minutes} & \textbf{Generations}\\
			\midrule
			\multirow{3}{*}{2} &
			A-$f_{U1}(c)$ & 0.5064 & 300 \\
			& A-$f_{U2}(c)$ & 0.2936 & 300 \\
			& A-$f_{U3}(c)$ & 0.6202 & 675 \\
			\hline
			\multirow{3}{*}{1} &
			B-$f_{U1}(c)$ & 63.1515 & 300 \\
			& B-$f_{U2}(c)$ & 63.4025 & 300 \\
			& B-$f_{U3}(c)$ & 103.1066 & 675 \\
			\hline
			\multirow{3}{*}{2} &
			GEN-$f_{U1}(c)$	& 0.5361 & 300 \\
			& GEN-$f_{U2}(c)$ & 0.6630 & 300 \\
			& GEN-$f_{U3}(c)$ & 0.9489 & 675 \\
			\bottomrule
		\end{tabular}%
		%}
		\label{tab:time_unsupervised}%
	\end{table}%
	
	\noindent{Even in the unsupervised case, the GA computational complexity on high dimensional data and with a wrapper approach can be observed. However, the computation time decreases significantly in respect to the one reported for the supervised approach, which is burdened by the 10-fold cross validation process.} \\
	The procedures executed on data matrices A and GEN require less then a minute to obtain the best fitness value, according to our stopping criteria. The maximum number of generation is increased, especially for the NFF experiments and the maximum time parameter constraint is never met.\\
	The experiments on data matrix B reveal to be faster then their supervised counterpart, never exceeding 103 minutes of computation. \\
	Therefore, the unsupervised approach may be valued as a good instrument to perform a preliminary analysis on the data matrices, without introducing any type of constraint.

	% % % % % DISCUSSION % % % % %
	%\section{Discussion}\label{sec:discussion}
	% This should explore the significance of the results of the work, not repeat them. A combined Results and Discussion section is often appropriate. Avoid extensive citations and discussion of published literature.

	% % % % % CONCLUSIONS % % % % %
	\section{Conclusions}\label{sec:conclusions} 
	%Riprendere i punti presi nell'intro. Enfasi sul risultato supervised/GEN e sul clustering che può essere utile per dare una lettura del dato.
	%Future work: relazioni spaziali, provare le feature selezionate dal clustering sul supervised, per sfruttare il tempo computazionale ed evitare il problema train/test
	% Commento sulle data matrix grandi, che possono essere tali anche perchè non introduciamo alcuna conoscenza a priori.
	In this study, the proposed approach for feature selection has revealed to be effective on heterogeneous groups of EEG data, considering the EEG signal variability in a single subject and in a group of subjects, different experimental paradigms and unbalanced task instances. In fact, two publicly available datasets on cognitive workload and motor movement/imagery have been used and combined to evaluate the generalizability of our proposal. The computation of various features in the time, frequency and time-frequency domain has provided a set of high dimensional data matrices, whose information are not too influenced by a priori knowledge. \\
	Our GA for feature selection has provided solutions that significantly reduce the feature number and performances that exceed or are comparable to the ones obtained by the benchmarking techniques. In particular, the performances obtained in the supervised approach, testing GA with our novel fitness function on the GEN data matrix, have significantly increased compared to the benchmark. Therefore, these reveals the effectiveness of our proposal on heterogeneous data and on more complex classification tasks. \\
	The proposed unsupervised approach has proved to be particularly efficient in terms of computational time and silhouette improvement, especially when coupled with our novel fitness function. The obtained results suggest that this approach could be employed to provide a completely data-driven preliminary assessment of the data matrix information.
	Moreover, it could be exploited to rapidly provide a possibly unbiased subset of features to pass as input of a supervised model.\\
	Another future development of this hybrid application of supervised and unsupervised approaches, may benefit from the introduction of expert knowledge to provide more balanced results on the silhouette score. This could be achieved by considering not only the silhouette score per se, but also the number of elements in each cluster, knowing the labels of each item.\\
	Finally, the a priori knowledge may be exploited to consider spatial relationships between the features, being them computed on specific electrodes deputed to the recording of specific cortical areas brain potentials.

	% % % % % REFERENCES % % % % %
	\bibliography{mybibfile}

\end{document}